\documentclass{article}

\PassOptionsToPackage{numbers, compress}{natbib}

 \usepackage[preprint]{neurips_2018}

\usepackage{lscape} 

\usepackage{graphicx}
\usepackage{subfigure}

\usepackage{graphicx} 
\usepackage{wrapfig} 
\usepackage{float}	  
\usepackage{calc} %
\usepackage{caption}
\usepackage{natbib} 

\usepackage{fontawesome} 
\usepackage[nolist,smaller]{acronym}

\usepackage{algorithm}
\usepackage[noend]{algpseudocode}

\usepackage{amsfonts, amsmath, amsthm} 
\usepackage{cancel} 
\usepackage{bbm, bm} 
\usepackage{IEEEtrantools}
\theoremstyle{definition}
\newtheorem{theorem}{Theorem}

\usepackage[dvipsnames]{xcolor}

\definecolor{navy}{RGB}{0, 0, 128}
\definecolor{pythonPurple}{RGB}{148,0,211}
\definecolor{darkgreen}{rgb}{0,0.3,0}
\definecolor{pythonOrange}{RGB}{255,165,0}

\definecolor{CBred}{RGB}{213, 94, 0}
\definecolor{CBblue}{RGB}{0,114,178}

\usepackage{tikz} 
\usetikzlibrary{arrows,shapes}
\usepackage{bigints} 

\usepackage[utf8]{inputenc} 
\usepackage[T1]{fontenc}    
\usepackage{hyperref}       
\usepackage{url}            
\usepackage{booktabs}       
\usepackage{amsfonts}       
\usepackage{nicefrac}       
\usepackage{microtype}      
\usepackage{xspace}

\usepackage{xcolor} 
\usepackage[utf8]{inputenc}
\usepackage[english]{babel}
\usepackage{amsthm}
\usepackage{multicol}

\newcommand{\acro}[1]{\textsc{#1}\xspace}

\newcommand{\DGP}{\acro{\smaller DGP}}
\newcommand{\DGPs}{\acro{\smaller DGPs}}
\newcommand{\GP}{\acro{\smaller GP}}

\newcommand{\GVI}{\acro{\smaller GVI}}
\newcommand{\ELBO}{\acro{\smaller ELBO}}
\newcommand{\KLD}{\acro{\smaller KLD}}

\newcommand{\BD}{\acro{\smaller $D_{B}^{(\beta)}$}}
\newcommand{\GD}{\acro{\smaller $D_{G}^{(\gamma)}$}}
\newcommand{\RAD}{\acro{\smaller $D_{AR}^{(\alpha)}$}}

\newcommand{\VI}{\acro{\smaller VI}}

\newcommand{\RMSE}{\acro{\smaller RMSE}}

\newcommand{\K}{\acro{\smaller K}}

\newcommand{\Lb}{\acro{\smaller $\mathcal{L}_{p}^{\beta}$}}
\newcommand{\Lg}{\acro{\smaller $\mathcal{L}_{p}^{\gamma}$}}

\DeclareMathOperator*{\argmin}{arg\,min}

\newcommand{\EPSRC}{\acro{\smaller EPSRC}}

\def\*#1{\boldsymbol{#1}} 
\definecolor{VIColor}{HTML}{D55E00}
\definecolor{GVIColor1}{HTML}{0072B2}
\definecolor{GVIColor2}{HTML}{56B4E9}
\definecolor{FVIColor}{HTML}{009E73}

\title{Robust Deep Gaussian Processes}

\author{
  Jeremias Knoblauch\\
  The Alan Turing Institute\\
  Dept. of Statistics\\
  University of Warwick\\
  \texttt{j.knoblauch@warwick.ac.uk} \\
}

\begin{document}

\maketitle

\begin{abstract} 
    This report provides an in-depth overview over the implications and novelty Generalized Variational Inference (\GVI) \citep{GVI} brings to Deep Gaussian Processes (\DGPs) \citep{DGPs}\footnote{If you cite this report, it is appropriate to cite these two papers, too.}.  
    Specifically, robustness to model misspecification as well as principled alternatives for uncertainty quantification are motivated with an information-geometric view.
    These modifications have clear interpretations and can be implemented in less than 100 lines of Python code.
    Most importantly, the corresponding empirical results show that \DGPs can greatly benefit from the presented enhancements.
\end{abstract} 

\section{Introduction}

Deep Gaussian Processes (\DGPs) were introduced by \citet{DGPs} and extend the logic of deep learning to the nonparametric Bayesian setting. 
The principal idea is to iteratively place Gaussian Process (\GP) priors over emerging latent spaces. 
More specifically, given observations $(\*X, \*Y)$ where $\*X \in \mathbb{R}^{n\times D}$ and $\*Y \in \mathbb{R}^{n \times p}$, a \DGP of $L$ layers introduces the additional collection of latent functions $\{\*F^l\}_{l=1}^L$.
Here, $\*F^l$ is a matrix of dimension $D^l\times D^{l+1}$.
Setting $\*F^0 = \*X$, $D^0 = D$ and $D^{l+1} = p$ for notational convenience, one can now write the hierarchical \DGP construction as
\begin{IEEEeqnarray}{rClCl}
    \*Y|\*F^L &&& \sim & 
    p\left( \*Y \middle|\; \*F^L \right)
    \nonumber \\
    \*F^L|\*F^{L-1} & = & f^{L}(\*F^{L-1}) & \sim & \GP\left(\mu^{L}(\*F^{L-1}), \K^{L}(\*F^{L-1}, \*F^{L-1}) \right) \nonumber \\
    \*F^{L-1}|\*F^{L-2} & = & f^{L-1}(\*F^{L-2})& \sim & \GP\left(\mu^{L-1}(\*F^{L-2}), \K^{L-1}(\*F^{L-2}, \*F^{L-2}) \right) \nonumber \\
         &&& \quad \dots \nonumber \\
    \*F^1|\*F^0 & = & f^{1}(\*F^{0})& \sim & \GP\left(\mu^1(\*F^0), \K^1(\*F^0, \*F^0) \right), \nonumber
\end{IEEEeqnarray}
where the mean and covariance functions are of form $\mu^l:\mathbb{R}^{D^l}\to\mathbb{R}^{D^{l+1}}$ and $\K^l:\mathbb{R}^{D^l\times D^l}\to\mathbb{R}^{D^{l+1}\times D^{l+1}}$.
Scalable inference in this construction is obviously a challenge. 
In principle, the attempts at tackling this problem  rely on Variational inference (\VI) strategies \citep{DGPs, VAEDGPs, DeepGPsVI, NestedVariationalDGPs}, Monte Carlo methods  \citep{SamplingDGPs, SMCDGPs} or more specialized approaches \citep{DGPEP, RandomFourierDGP}.
In the remainder, we will focus on \VI strategies for \DGP inference. To keep things as simple as possible, we discuss the implications of Generalized Variational Inference (\GVI) only in relation to the arguably most promising \VI approach of \citet{DeepGPsVI} which encodes conditional dependence into the variational family $\mathcal{Q}$.
We note in passing however that due to \GVI's versatility, the same logic applies to any other variational family $\mathcal{Q}$.

The rest of this report is structured as follows: First, we give a brief recap of the variational families for \DGPs proposed by \citet{DeepGPsVI}. Next, we give a brief overview of \GVI as introduced in \citet{GVI}. Lastly, we give the necessary derivations necessary to apply \GVI to \DGPs and investigate the performance gains.

\section{Variational inference  for Deep Gaussian Processes}

This report focuses on the variational family $\mathcal{Q}$ for \DGPs geared towards large-scale inference introduced by \citet{DeepGPsVI}.
Unlike competing \VI approaches for \DGPs, this family encodes some part of the conditional dependence structure of the \DGP. 
This comes at the expense of losing a tractable closed form lower bound \citep[as in][]{DGPs}, but makes \DGPs more flexible and adaptable.

\subsection{The conditionally dependent variational family for \DGPs}
Including the inducing point framework for \GP inference \citep[see][]{InducingOriginal, InducingBonilla, InducingProcesses}, we now introduce the exact Bayesian posterior arising from the \DGP construction.
First, define the set of $m$ additional inducing points $\*Z^l = (\*z^{l}_1, \*z^{l}_2, \dots, \*z^{l}_m)^T$ and their function values $\*U^l = (f^l(\*z^{l}_1), f^l(\*z^{l}_2), \dots, f^l(\*z^{l}_m))^T$. 
For better readability, we will often drop $\*X$ and $\*Z^l$ from the conditioning sets.
Further, note that we will denote the $i$-th row of the $D^l \times D^{l+1}$ latent functions $\*F^l$ as $\*f^{L}_i$.
With this in place, the joint distribution of the \DGP construction is
\begin{IEEEeqnarray}{rCl}
    p\left(\*Y, 
    \{\*F^l\}_{l=1}^L, \{\*U^l\}_{l=1}^L
    \right)
    & = &
    \underbrace{\prod_{i=1}^n p(\*y_i|\*f^{L}_i)}_{\text{likelihood}} \times
    \underbrace{\prod_{l=1}^L 
        p\left(\*F^l \middle|\; \*U^l, \*F^{l-1}, \*Z^{l-1} \right)p\left( \*U^l \middle|\; \*Z^{l-1}  \right)}_{\text{ (\DGP) prior}}. \quad\quad \nonumber
\end{IEEEeqnarray}
Thus, the posteriors $p\left( \{\*F^l\}_{l=1}^L, \{\*U^l\}_{l=1}^L
\right)$ and $p\left( \{\*F^l\}_{l=1}^L 
\right)$ 
%
are intractable due to the required normalizing constants required for their computation.
To overcome this, different variational approximations have been proposed.
Here, we focus on the variational family proposed in \citet{DeepGPsVI} given by
\begin{IEEEeqnarray}{rCl}
    q\left( \{\*F^l\}_{l=1}^L, \{\*U^l\}_{l=1}^L 
    \right) 
    & = & 
    \prod_{l=1}^L 
        p\left(\*F^l \middle|\; \*U^l, \*F^{l-1}, \*Z^{l-1} \right)
        q\left( \*U^l \right), 
        \label{eq:variationalFam1}
        \\
    q\left( \*U^l \right)
    & = &
        \mathcal{N}\left(\*U^l\middle|\; \*m^l, \*S_l \right).
    \label{eq:variationalFam2}
\end{IEEEeqnarray}
The collection of variational parameters for this posterior is given by $\left\{ \{\*m^l\}_{l=1}^L, \{\*S_l\}_{l=1}^L  \right\}$. 
The normal form for $q\left( \*U^l \right)$ is chosen because it allows for exact integration over the inducing points $\{\*U^l\}_{l=1}^L$, yielding the closed form variational posterior  
\begin{IEEEeqnarray}{rCl}
    q\left(\{\*F^l\}_{l=1}^L
    \right)
    & = &
    \prod_{l=1}^L
    \mathcal{N}\left(\*F^l \middle|\; \*\mu^l, \*\Sigma_l \right),
\end{IEEEeqnarray}
where the parameters of the posterior are available in closed form as
\begin{IEEEeqnarray}{rClCl}
    \left[\*\mu^l\right]_{i} & = &
    \*\mu^{\*m^l, \*Z^{l-1}}(\*f^{L}_i) & = &
    \mu^l(\*f^{L}_i) + \*a(\*f^{L}_i)^T\left(\*m^l - \mu^l(\*Z^{l-1})\right)\\
    \left[\*\Sigma_l\right]_{i,j} & = &
    \*\Sigma_{\*S_l, \*Z^{l-1}}(\*f^{L}_i, \*F^{j,l}) & = & \K^l(\*f^{L}_i, \*F^{j,l}) - \*a(\*f^{L}_i)^T\left(\K^l(\*Z^{l-1}, \*Z^{l-1}) - \*S_l\right)\*a(\*F^{j,l}),
\end{IEEEeqnarray}
%
where as usual we define $\*a(\*f^{L}_i) = \K^l(\*Z^{l-1}, \*Z^{l-1})^{-1}\K^l(\*Z^{l-1}, \*f^{L}_i)$.
Note the attractive feature of the family specified via eqs. \eqref{eq:variationalFam1} -- \eqref{eq:variationalFam2}: At each layer $l$, the output $\*f^{l}_i$ only depends on the corresponding input $\*f^{l-1}_i$. 
This property is a direct consequence of setting every layer up exactly as a sparse \GP \citep[see, e.g.][]{InducingOriginal, GPforBigData, InducingBonilla}.
This enables efficient probabilistic backpropagation \citep{PBP} with the reparameterization trick \citep[e.g.][]{gradTrick, reparameterizationTrick} and makes the approach scalable.

In particular, \citet{DeepGPsVI} propose a doubly stochastic minimization of the negative Evidence Lower Bound (\ELBO) given by 
\begin{IEEEeqnarray}{rCl}
    \mathcal{L}(q|\*Y, \*X) & = &
    -\sum_{i=1}^n\mathbb{E}_{q(\*f^{L}_i)}
    \left[ 
        \log p(\*y_i|\*f^{L}_i)
    \right]
    +
    \sum_{l=1}^L\KLD(q(\*F^l, \*U^l)||p(\*F^l, \*U^l |\*Z^{l-1})).
    \label{eq:DGP_elbo_old}
\end{IEEEeqnarray}
The Kullback-Leibler divergence  (\KLD) terms of this bound further simplify  because by eq. \eqref{eq:variationalFam1}, $q$ is designed to cancel the conditional over $\*F^l$ with $p$. This finally leads to the bound
\begin{IEEEeqnarray}{rCl}
    \mathcal{L}(q|\*Y, \*X) & = &
    -\sum_{i=1}^n\mathbb{E}_{q(\*f^{L}_i)}
    \left[ 
        \log p(\*y_i|\*f^{L}_i)
    \right]
    +
    \sum_{l=1}^L\KLD(q(\*U^l)||p(\*U^l|\*Z^{l-1})),
    \label{eq:DGP_elbo}
\end{IEEEeqnarray}
where for optimization the samples for $\*F^l$ are drawn using the variational posteriors from the previous layers. Because $\*f^{L}_i$ only depends on the corresponding input $\*F^{i, l-1}$, this can be done using univariate Gaussians and thus does not involve matrix operations.
Approximating the expectation over $q(\*\theta)$ induces the first layer of stochasticity in this model.
The second layer is due to drawing a mini-batches from $\*X = \*F^0$ and $\*Y$ at each iteration.
Because of this degree of stochasticity, it is an appealing feature that the expectations $\mathbb{E}_{q(\*f^{L}_i)}\left[\log p(\*y_i|\*f^{L}_i)\right]$ are available in closed form for some choices of $p$. 
This is for instance the case for the regression setting, where $p$ is a normal likelihood. 
Later on, we also derive such closed forms for a new class of alternatives for $p$ geared towards robustness and derived from normal likelihoods.

\subsection{An alternative problem representation}
We now decompose the components of the \DGP model. Specifically, we define the collection of likelihood terms as 
\begin{IEEEeqnarray}{rCl}
    \ell_n\left(\{ \{\*F^l\}_{l=1}^L,  \{\*U^l\}_{l=1}^L\}, \*Y\right) &=& \sum_{i=1}^n\ell\left(\*f^{L}_i, \*y_i \right) \;\; \text{ for }  \ell\left(\*f^{L}_i, \*y_i \right) =  -\log p(\*y_i|\*f^{L}_i) 
\end{IEEEeqnarray}
and the layered \DGP prior via
\begin{IEEEeqnarray}{rCl}
    p\left(\{\*F^l\}_{l=1}^L, \{\*U^l\}_{l=1}^L \middle|\; \{ \*Z^{l} \}_{l=1}^L \right) &= & \prod_{l=1}^L 
        p_l\left(\*F^l, \*U^l \middle|\; \*F^{l-1}, \*U^{l-1}, \*Z^{l-1}  \right)
        \label{eq:DGP_prior} \\
    p_l\left(\*F^l, \*U^l \middle|\; \*F^{l-1}, \*U^{l-1}, \*Z^{l-1}  \right) & = &
        p\left(\*F^l \middle|\; \*F^{l-1}, \*U^l, \*Z^{l-1} \right)p\left( \*U^l \middle|\; \*Z^{l-1}  \right).
        \label{eq:DGP_prior_l}
\end{IEEEeqnarray}
With this, one can rewrite the sought-after posterior as 
\begin{IEEEeqnarray}{rCl}
    &&p\left( 
    \{\*F^l\}_{l=1}^L, \{\*U^l\}_{l=1}^L
    \middle|\; 
    \*Y, \*X \right)
    \nonumber \\
     &=& \dfrac{
    \exp\left\{ 
        -\ell_n\left(\{ \{\*F^l\}_{l=1}^L,  \{\*U^l\}_{l=1}^L\}, \*Y\right)
        \right\}   
    \pi\left(
        \{\*F^l\}_{l=1}^L, \{\*U^l\}_{l=1}^L \middle|\; \{ \*Z^{l} \}_{l=1}^L 
    \right)}{
        \int_{\*Y}\exp\left\{ 
        -\ell_n\left(\{ \{\*F^l\}_{l=1}^L,  \{\*U^l\}_{l=1}^L\}, \*Y\right)
        \right\}   
    \pi\left(
        \{\*F^l\}_{l=1}^L, \{\*U^l\}_{l=1}^L \middle|\; \{ \*Z^{l} \}_{l=1}^L 
    \right)d\*Y
    } 
    \label{eq:DGP_gen_posterior}
\end{IEEEeqnarray}
This representation gives a generalized Bayesian distribution 
associated with a general loss function $\ell$. 
For the standard \DGP, the loss function is the negative log likelihood $\ell(\*f^{L}_i, \*y_i) = -\log p(\*y_i|\*f^{L}_i)$, which is the loss traditionally associated with the Bayesian paradigm. 
As part of this report, we explain alternative losses $\ell_n$ for the probabilistic \DGP model \citep[as in][]{GVI}. 
However, unlike the log likelihood these losses will be robust to model misspecification and outliers.
Note that the variational methods outlined in the previous section still apply to any new additive loss $\ell$. In fact, one only needs to replace $-\log(p(\*y_i|\*f^{L}_i))$ in eq. \eqref{eq:DGP_elbo} with the alternative loss $\ell(\*f^{L}_i, \*y_i)$.
%

\section{Generalized Variational Inference}

Unlike \VI where the quality of the posterior is controlled only via the variational family $\mathcal{Q}$, Generalized Variational Inference (\GVI) allows for adapting two additional objects: The loss used for inference and the manner of uncertainty quantification.
For notational convenience, we formulate \GVI in full generality for a generic parameter $\*\theta$ of interest for inference. 
For the purposes of this report, this parameter indexes a \DGP and is $\*\theta = \{ \{\*F^l\}_{l=1}^L, \{\*U^l\}_{l=1}^L \}$.

In a nutshell, \GVI is the natural methodological outgrowth arising from the study of a generalized representation of Bayesian inference. This representation recovers standard Bayesian inference, \VI and many other methods as special cases.
Specifically, \citet{GVI} axiomatically derive Bayesian inference as the triplet $P(\ell_n, D, \Pi)$ given by
\begin{IEEEeqnarray}{rClCCC}
    q^{\ast}(\*\theta) & = & \argmin_{q \in \Pi} \left\{\mathcal{L}(q|\*Y, \*X, \ell_n, D)\right\}; \;\;\; 
    \mathcal{L}(q|\*Y, \*X, \ell_n, D) & = &
        \mathbb{E}_{q(\*\theta)}
        \left[
            \ell_n(\*\theta, \*Y)
        \right]  + D\left(q||p\right). \quad\quad
    \label{eq:GeneralGeneral_objective}
\end{IEEEeqnarray}
where $D\left(q||p\right)$ depends on $\*X$ for the \DGP. Denoting by $\mathcal{P}(\*\Theta)$ the space of all probability distributions over $\*\Theta$, the constituent parts 
of the form $P(\ell_n, D, \Pi)$ are given by\\[-0.45cm]
\begin{itemize}
    \item a \textbf{loss} $\ell_n$ linking a parameter of interest $\*\theta$ to the observations $\*Y = \*y_{1:n}$. This loss is assumed to be additive throughout, i.e. $\ell_n(\*\theta, \*Y) = \sum_{i=1}^n\ell(\*\theta, \*y_i)$ for some $\ell$. \\[-0.4cm]
    \item a {divergence ${D}:\mathcal{P}(\*\Theta)\times \mathcal{P}(\*\Theta) \to \mathbb{R}_+$ regularizing the posterior} with respect to the prior. As $D$ determines how the prior $p$ quantifies uncertainty, it is called \textbf{uncertainty quantifier}. 
    Note that for the \DGP, $\*X$ enters eq. \eqref{eq:GeneralGeneral_objective} via $D(q||p)$; \\[-0.4cm]
    \item a set of \textbf{admissible posteriors $\Pi \subseteq \mathcal{P}(\*\Theta)$} the regularized expected loss is minimized over. 
    \\[-0.6cm]
\end{itemize} 
The seminal paper of \citet{Zellner} shows that standard Bayesian inference solves $P(-\sum_{i=1}^n\log(p(\*\theta|\*y_i)), \KLD, \mathcal{P}(\*\Theta))$. 
This is extended in \citet{Bissiri}, who show that for an additive loss function $\ell_n$, the Gibbs-posterior is the solution to $P(-\sum_{i=1}^n\log(p(\*\theta|\*y_i)), \KLD, \mathcal{P}(\*\Theta))$.
Further, for $\mathcal{Q}$ a variational family, the objective of $P( -\sum_{i=1}^n\log(p(\*\theta|\*y_i)), \KLD, \mathcal{Q})$ in eq. \eqref{eq:GeneralGeneral_objective} is the Evidence Lower Bound (\ELBO) of \VI.
This observation is the inspiration to call any problem of form $P(\ell_n, D,  \mathcal{Q})$ a \textbf{Generalized Variational Inference (\GVI)} problem.
Perhaps the most interesting aspect of \GVI lies in its modularity. Roughly speaking, once $\mathcal{Q}$ is fixed, this modularity allows one to prove that (i) robustness to model misspecification should enter the Bayesian inference problem via $\ell_n$ and that (ii) a change in the posterior shape should enter via $D$ \citep[see Thm. 5 in][]{GVI}.

%

\section{Generalized Variational Inference for Deep Gaussian Processes}

With $\*\theta = \{ \{\*F^l\}_{l=1}^L, \{\*U^l\}_{l=1}^L \}$, it is clear that \eqref{eq:GeneralGeneral_objective} allows for tractable alternatives of eq. \eqref{eq:DGP_elbo}.
This section explains which roles $\ell_n$ and $D$ take in the \GVI formulation and derive appropriate choices for robust \DGPs. We do not discuss $\Pi$, since throughout, we focus on the case where $\Pi = \mathcal{Q}$ using the variational family $\mathcal{Q}$  of \citet{DeepGPsVI} introduced above.

\subsection{Model-agnostic and likelihood-based losses for robustness against misspecification}
In traditional Bayesian inference, the loss term $\ell_n$  is a sum over negative log likelihoods. 
Yet, this is just a special case \cite{Bissiri} and $\ell_n(\*\theta, \*Y)$ can be \textit{any} additive loss about whose optimum $\*\theta^{\ast}$ one wishes to learn in a Bayesian manner.
In fact, using the notation introduced above, \citet{Bissiri} show that for the exact Bayesian inference problem $P(\ell_n, \KLD, \mathcal{P}(\*\Theta))$, one recovers the generalized Bayes Theorem 
\begin{IEEEeqnarray}{rCl}
    q^{\ast}(\*\theta) & = &
        \dfrac{p(\*\theta)\exp
            \left\{ 
                \sum_{i=1}^n-\ell(\*\theta, \*y_i)
            \right\}
        }{
            \int_{\*\Theta}
            p(\*\theta)\exp
            \left\{ 
                \sum_{i=1}^n-\ell(\*\theta, \*y_i)
            \right\}d\*\theta
        },
        \label{eq:Generalized_Bayes_Thm}
\end{IEEEeqnarray}
which mirrors the form in eq. \eqref{eq:DGP_gen_posterior}.
We note that updating rules of this kind have been studied under the name of Gibbs- or Pseudo-posteriors before, but \citet{Bissiri} show that they are indeed valid and coherent posterior beliefs about $\*\theta$ in their own right.

Inspired by this insight, various authors have proposed likelihood-based losses replacing the negative log likelihood $-\log p(\*\theta, \*y_i)$ but enabling inference \textit{in the same model} described by the likelihood $p$ and the same parameter $\*\theta$. 
Usually, this is done for robust inference and a recent overview is provided for in \citet{Jewson}.
The recipe for deriving these alternative losses derives from geometric considerations. 
In particular, when minimizing $\sum_{i=1}^n-\log p(\*y_i|\*\theta)$ to conduct inference on $\*\theta$, one implicitly minimizes the (non-robust) \KLD in the space of densities.
To see that this is the case, denote by $g$ the data-generating probability density (i.e., $\*y_1, \*y_2, \dots, \*y_n \sim g$) and observe that
\begin{IEEEeqnarray}{rCl}
    \frac{1}{n}\sum_{i=1}^n-\log(p(\*y_i|\*\theta) \approx \mathbb{E}_{g}[-\log(p(\*y|\*\theta))] = \KLD(g||p(\cdot|\*\theta)) + \mathbb{E}_{g}[\log(g(\*y))].
    \label{eq:KLD^loss}
\end{IEEEeqnarray}
Since the entropy term $\mathbb{E}_{g}[\log(g(\*y))]$ does not depend on $\*\theta$, minimizing the negative log likelihood thus amounts to (approximately) minimizing the \KLD between the true data generating mechanism and the model as parameterized by $\*\theta$. 
While the \KLD is a good measure of discrepancy if the model is an appropriate description for the data-generating mechanism, this no longer holds under moderate model misspecification or outliers.
In traditional statistical inference, this is usually not a problem: A lot of effort is typically  expanded in order to investigate the data patterns and adapt $p$ to be a better description of $g$. 
In modern statistical machine learning and its accompanying black box methods and variational approximations, this is no longer the case: Moderate, even severe misspecification is the norm.
%


\begin{figure}[t!]
\vskip -1.0cm
\begin{center}
\includegraphics[trim= {0.0cm 0.5cm 0.9cm 0.2cm}, clip,  
width=0.49\columnwidth]{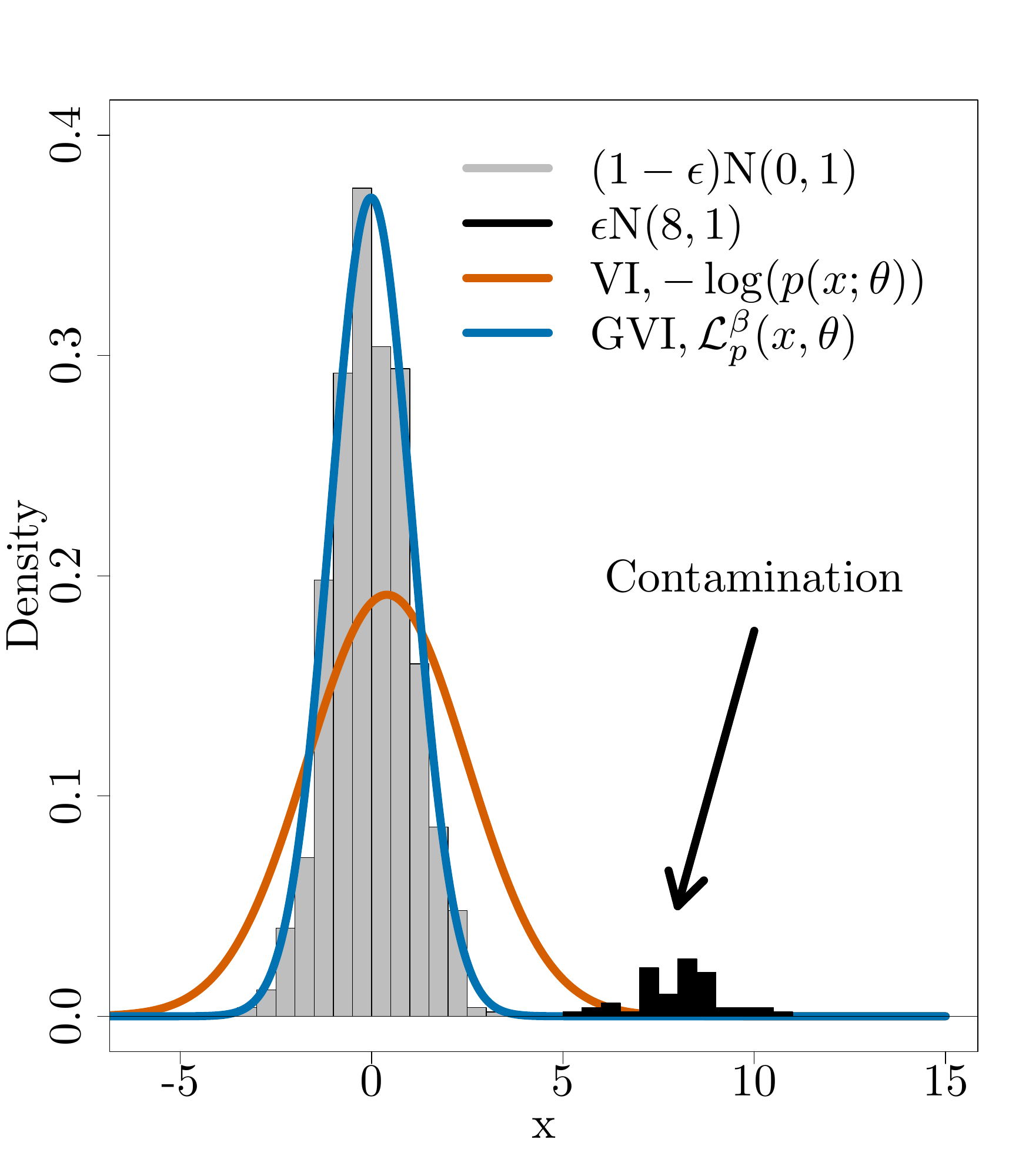}
\includegraphics[trim= {0.0cm 0.5cm 0.9cm 0.2cm}, clip,  
width=0.49\columnwidth]{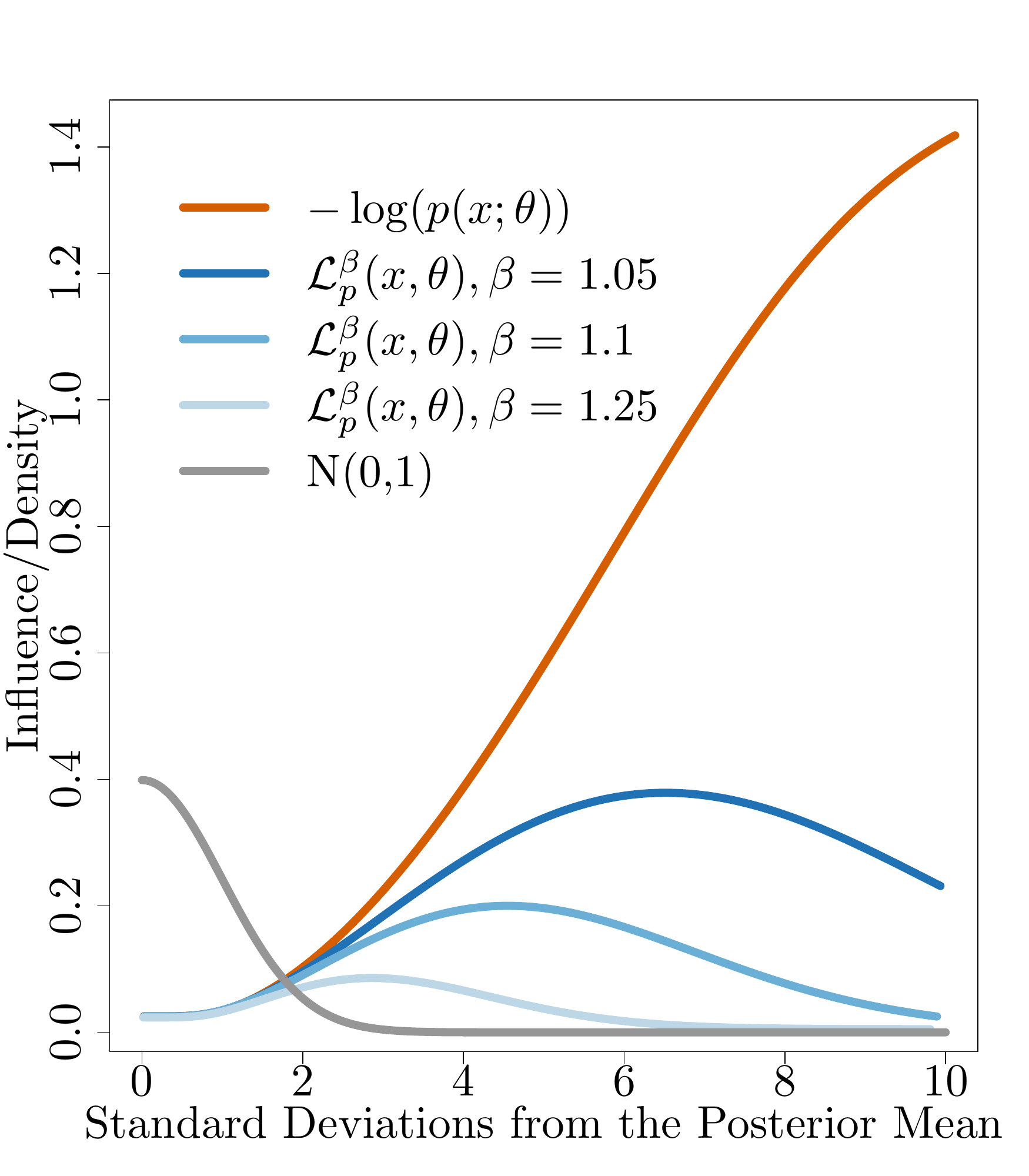}
\caption{
    Taken from \citep{RBOCPD, GVI}.
    Comparing likelihood-based robust losses usable within \textcolor{GVIColor1}{\textbf{\GVI}} with the standard negative log likelihood loss used within \textcolor{VIColor}{\textbf{\VI}}.
    \textbf{Left}: 
    Transforming the loss provides robustness against model misspecification. Depicted are
    posterior predictives under $\varepsilon = 5\%$ outlier contamination using 
     \textcolor{VIColor}{\textbf{\VI}}
    and  \textcolor{GVIColor1}{$P(\sum_{i=1}^n\Lb(\*\theta, \*y_i),\KLD, \mathcal{Q})$}, with $\Lb(\*\theta, \*y_i)$
    as in eq. \eqref{eq:BD^loss} for $\beta=1.5$. %
    \textbf{Right:}
    Depicted is the influence \citep[see][]{InfFct} of the 100th observation $\*y_{100}$ on exact posteriors for robust and non-robust losses in standard deviations from the posterior mean after 99 observations. 
    Higher influence is assigned for \textcolor{VIColor}{$\mathbf{-\log(p(\*y_i, \*\theta))}$} the more unlikely $\*y_i$ is under the current model.
    In contrast, \textcolor{GVIColor1}{$\mathbf{\Lb(\*\theta, \*y_i)}$} guards against assigning the highest influence to outliers.
    $\Lg(\*\theta, \*y_i)$ behaves similarly.
    %
    %
}
\label{Fig:robust_losses}
\vskip -0.25in
\end{center}
\end{figure}

In these circumstances, using the \KLD as a measure of discrepancy can often adversely affect inference outcomes. 
An information-geometrically elegant solution to this conundrum is changing the measure of discrepancy in the space of probability measures.
Numerous authors have pursued this idea \citep[see e.g.][]{MinDisparities, BasuDPD, GammaDivSummable, GammaDivNotSummable, Jewson}. 
For the robust \DGP, we focus on the $\beta$-divergences (\BD) and $\gamma$-divergences (\GD) \citep[see also][]{GoshBasuPseudoPosterior, Jewson, RBOCPD, AISTATSBetaDiv}. 
The logic is analogous to eq. \eqref{eq:KLD^loss} and leads to alternative loss functions. 
For example, the $\beta$-divergence is given by
\begin{IEEEeqnarray}{rCl}
\BD(g||p(\cdot |\*\theta))
&=&\frac{1}{\beta(\beta-1)}
    \mathbb{E}_g\left[ g(\*y)^{\beta-1}\right]
    +\frac{1}{\beta}
    \int p(\*y|\*\theta)^{\beta}d\*y
    -\frac{1}{\beta-1}
    \mathbb{E}_g\left[p(\*y|\*\theta)^{\beta-1}\right],\nonumber
\end{IEEEeqnarray}
%
and it is obvious that the first term does not depend on $\*\theta$. Thus, using the natural approximation of the expectation over $g$, one can target this divergence via $\frac{1}{n}\sum_{i=1}^n\Lb(\*\theta, \*y_i)$, where
%
%
\begin{IEEEeqnarray}{rCl}
    \Lb(\*\theta, \*y_i) & = &
        -
			\frac{1}{\beta-1}
			p(\*y_i|\*\theta)^{\beta-1} 
			+ \frac{I_{p, \beta}(\*\theta)}{\beta}
		\label{eq:BD^loss}
\end{IEEEeqnarray}
and $I_{p, c}(\*\theta) =  \int p(\*y|\*\theta)^{	c}d\*y$. 
The derivation for the $\gamma$-divergence is similar \citep[][]{GammaDivSummable} and yields
\begin{IEEEeqnarray}{rCl}
	\Lg(\*\theta, \*y_i) & = &
        -
			\frac{1}{\gamma-1}p(\*y_i|\*\theta)^{\gamma-1} \cdot
			  \frac{\gamma}{I_{p, \gamma}(\*\theta)^{-\frac{\gamma-1}{\gamma}}}
			. \; \label{eq:div_losses} 
\end{IEEEeqnarray}
For the \DGP, these losses simplify as $p(\*y_i|\*\theta) = p(\*y_i| \{\*F^l\}_{l=1}^L, \{\{\*U^l\}_{l=1}^L \}) = p(\*y_i|\*f^{L}_i)$. For clarity and brevity, this report thus uses $\Lb(\*\theta, \*y_i) = \Lb(\*f^{L}_i, \*y_i)$ and $\Lg(\*\theta, \*y_i) = \Lg(\*f^{L}_i, \*y_i)$. 

Through tedious but straightforward calculation, one can show that the corresponding expectations $\mathbb{E}_{q(\*f^{L}_i)}\left[\Lb(\*f^{L}_i, \*y_i)\right]$ and $\mathbb{E}_{q(\*f^{L}_i)}\left[\Lg(\*f^{L}_i, \*y_i)\right]$ are available in closed form for the regression setting where $p$ is a normal likelihood \citep[see supplementary material of][]{GVI}.

\begin{theorem}[Closed form for robust regression]
    If it holds that $\*y_i \in \mathbb{R}^d$,
    \begin{IEEEeqnarray}{rClCrCl}
        p(\*y_i|\*f^{L}_i) & = & \mathcal{N}\left(\*y_i; \*f^{L}_i, \sigma^2I_d\right); & \quad &
        q(\*f^{L}_i) & = & \mathcal{N}(\*f^{L}_i; \*\mu, \*\Sigma), 
    \end{IEEEeqnarray} 
    then for the quantities given by
    \begin{IEEEeqnarray}{rClCrClCrCl}
        \widetilde{\*\Sigma}^{-1} & = & \left(\frac{c}{\sigma^s}\*I_d + \*\Sigma^{-1}\right);
        & \quad &
        \widetilde{\*\mu} & = & \left( \frac{c}{\sigma^2}\*y_i + \*\Sigma^{-1}\*\mu \right);
        & \quad &
        I(c) & = & (2\pi\sigma^2)^{-0.5dc}c^{-0.5d}
    \end{IEEEeqnarray}
    and for 
    \begin{IEEEeqnarray}{rCl}
        E(c) & = & \frac{1}{c}
    \left({2\pi}\sigma^2\right)^{-0.5dc}
    \frac{|\widetilde{\*\Sigma}|^{0.5}}{ |\*\Sigma|^{0.5} }
    \exp\left\{
 -\frac{1}{2}\left(
     \frac{c}{\sigma^2}\*y_i^T\*y_i + \*\mu^T\*\Sigma^{-1}\*\mu -
     \widetilde{\*\mu}^T\widetilde{\*\Sigma}\widetilde{\*\mu}
 \right)
    \right\} 
    \end{IEEEeqnarray}
    the following expectations are available in closed form:
    \begin{IEEEeqnarray}{rCl}
        \mathbb{E}_{q(\*f^{L}_i)}\left[\Lb(\*f^{L}_i, \*y_i)\right] & = &
        -E(\beta-1) + \frac{I(\beta)}{\beta}
        \\
        \mathbb{E}_{q(\*f^{L}_i)}\left[\Lb(\*f^{L}_i, \*y_i)\right] & = &
        -E(\gamma-1)\cdot\frac{\gamma}{I(\gamma)^{\frac{\gamma}{\gamma-1}}}
    \end{IEEEeqnarray}
    \label{Thm:closedFormExp}
\end{theorem}

Fig. \ref{Fig:robust_losses} demonstrates that misspecification can be a severe detriment for inference with the negative log likelihood. It also uses influence functions to showcase how the alternative model-based losses $\Lb$ and $\Lg$ can avoid suffering under model misspecification.
We note in passing that for numerical stability, \Lg is the preferable loss since it is multiplicative and unlike \Lb never changes sign. Thus, it can be processed and stored entirely in log form.

\subsection{Alternative uncertainty quantification for prior robustness and marginal variances} 
\begin{figure}[t!]
\vskip -1cm 
\begin{center}
\includegraphics[trim= {0.0cm 0.5cm 0.9cm 0.2cm}, clip,  
width=0.49\columnwidth]{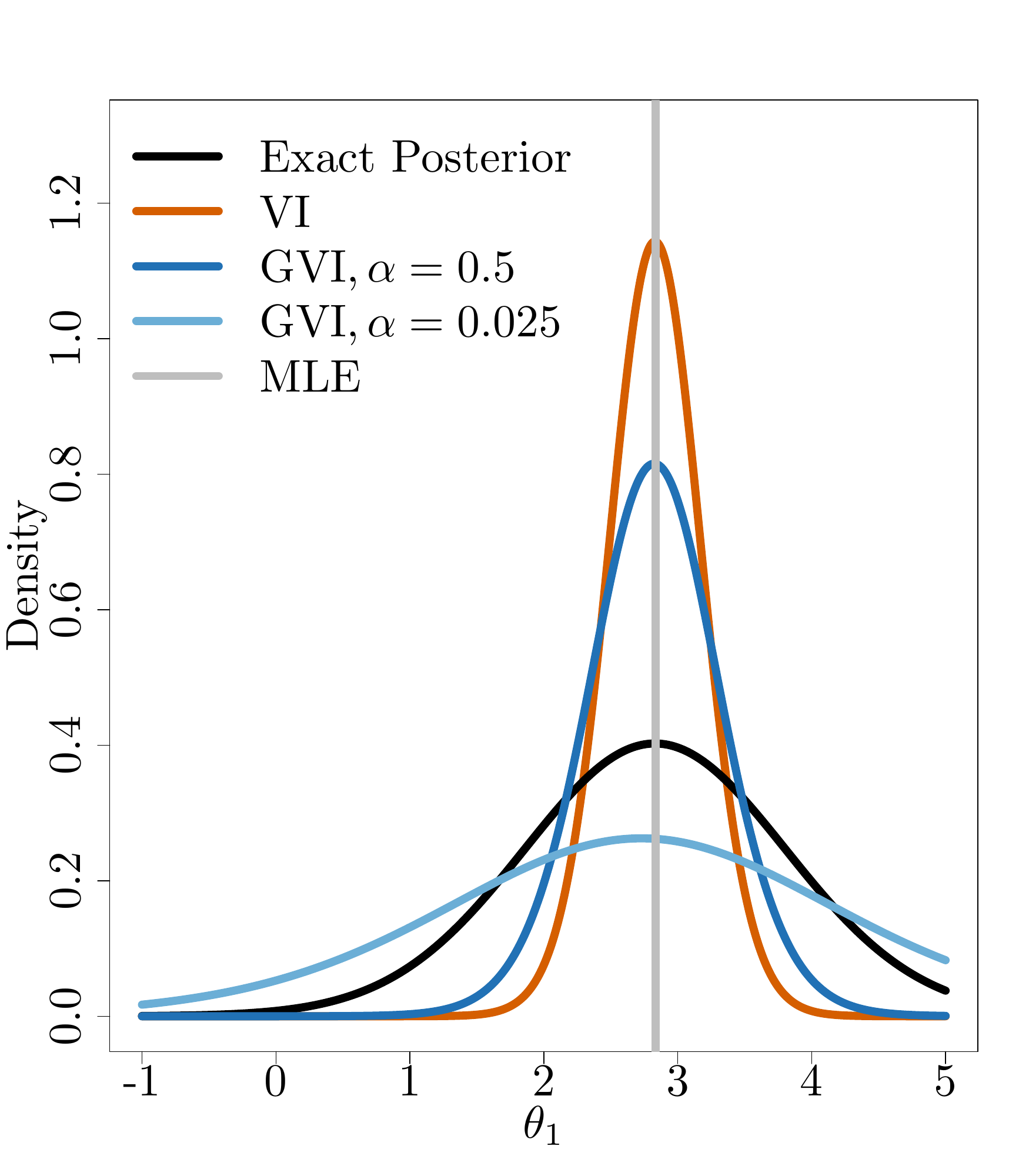}
\includegraphics[trim= {0.0cm 0.5cm 0.9cm 0.2cm}, clip,  
width=0.49\columnwidth]{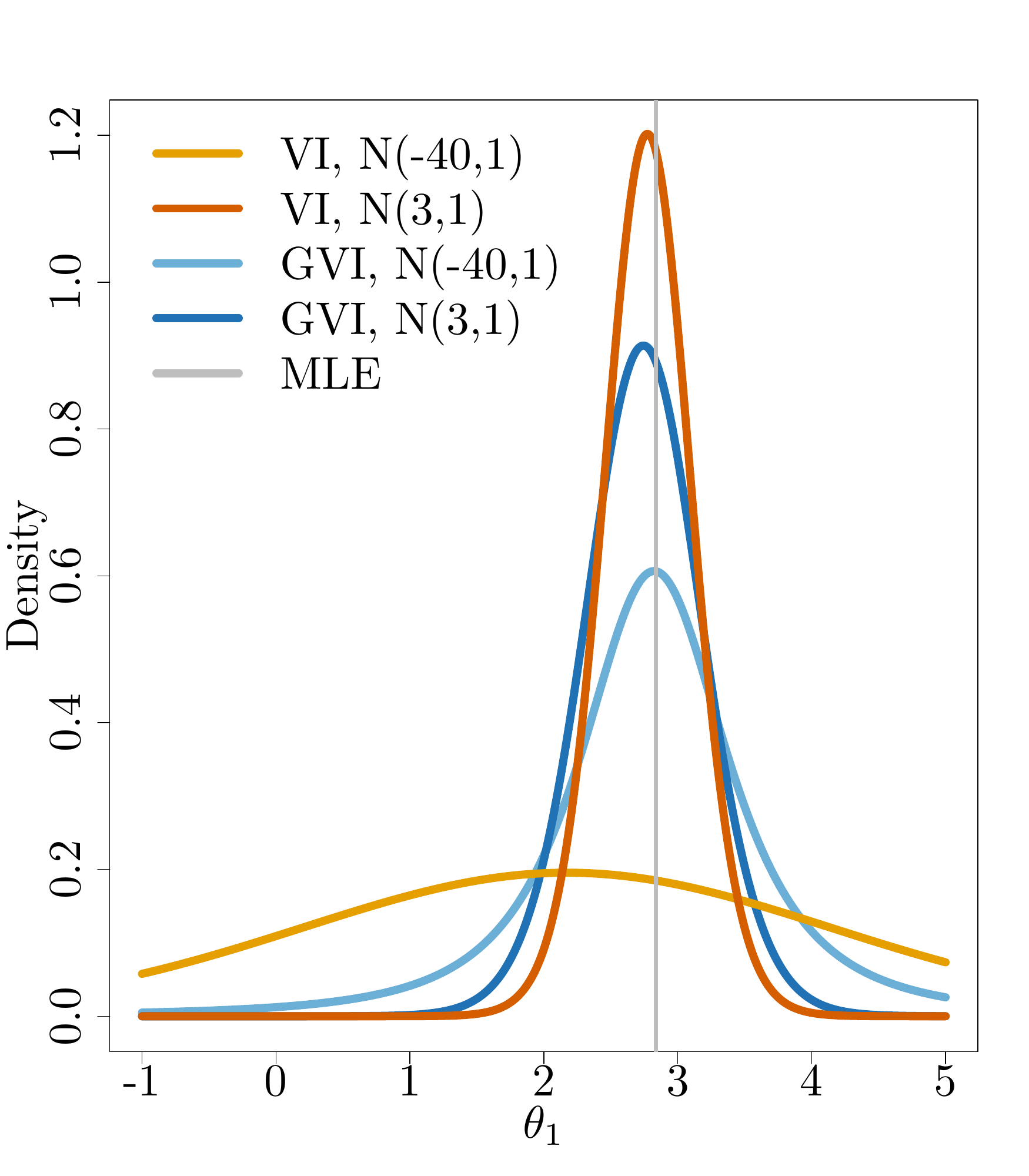}
\caption{
Taken from \citep{GVI}.
Comparing standard \textcolor{VIColor}{\textbf{\VI}} ($D= \KLD$) against \textcolor{GVIColor1}{\textbf{\GVI}} with $D = \RAD$ using posteriors with Gaussian likelihoods and mean-field Gaussian approximations. 
\textbf{Left:}
Changing $D$ improves marginal variances.
%
Depicted are exact and approximate marginals. The exact posterior is correlated, causing \textcolor{VIColor}{\textbf{\VI}} to over-concentrate. \textcolor{GVIColor1}{\textbf{\GVI}} can avoid this.
\textbf{Right:} Changing $D$ provides prior robustness.
Depicted are approximate marginals for two different priors $\pi \in \{N(-30,2^2), N(-5,2^2)\}$. \textcolor{VIColor}{\textbf{\VI}} is sensitive to the badly specified prior. \textcolor{GVIColor1}{\textbf{\GVI}} can avoid this.
}
\label{Fig:alternative_UQ}
\vskip -0.25in
\end{center}
\end{figure}

In contrast to Maximum Likelihood inference, Bayesian methods provide uncertainty quantification about $\*\theta$. 
Specifically, uncertainty about $\*\theta$ is quantified by penalizing how far the posterior $q$ $D$-diverges from the prior $\pi$.
\GVI is the first method relaxing the constraint that $D = \KLD$.
Specifically, \citet{GVI} study robust alternatives to the \KLD. While \GVI is not limited to other divergences, in this report we focus on R\'enyi's $\alpha$-divergence \RAD (with the parameterization of \citet{ABCdiv}) given by
\begin{IEEEeqnarray}{rCl}
    \RAD(q(\*\theta)||p(\*\theta))&=&\frac{1}{\alpha(\alpha-1)}\log\left(\int q(\*\theta)^{\alpha}p(\*\theta)^{1-\alpha}d\theta+1\right).
\end{IEEEeqnarray}
This divergence is available in closed form for the variational families and priors on \DGPs of \citep{DeepGPsVI} for $\alpha \in (0,1)$. More importantly, it provides larger marginal variances than \VI for $\alpha \in (0,1)$, tighter marginal variances than \VI for $\alpha > 1$ and is robust to badly specified priors. We refer to Fig. \ref{Fig:alternative_UQ} for an illustration of both properties. 
We note that the supplementary material of \citep{GVI} contains a much wider selection of pictorial examples that also encompass other divergences. 
%

As a second alternative to \RAD-uncertainty quantification, we also consider $D = \frac{1}{w}\KLD$ \citep[see also][]{alpha-VI}. 
Note that this has an intimate relationship to power likelihoods. 
In particular, using the negative power log likelihood $-\log p(\*y_i|\*\theta)^w = -w\log p(\*y_i|\*\theta)$ as the loss in eq. \eqref{eq:GeneralGeneral_objective} gives the same solution as using the standard log likelihood together with $D = \frac{1}{w}\KLD$.
More generally and using the notation introduced above, $P(w\ell_n, D, \Pi) = P(\ell_n, \frac{1}{w}\KLD, \Pi)$.
We note in passing that for $w\in(0,1)$ this choice of $D$ places \textit{more} weight on the prior. Thus, contrary to \RAD it is anti-robust to the prior.
For $D=\KLD$, $D(q||p)$ has closed form if both $q$ and $p$ are (multivariate) normal densities. Next, we show that this discrepancy is available in closed form for $D = \RAD$, too \citep[see also][]{GVI}.

\begin{theorem}[closed forms for $D = \RAD$]
    For $q(\*\theta) = \mathcal{N}(\*\theta; \*\mu^q, \*\Sigma_q)$ and $p(\*\theta) = \mathcal{N}(\*\theta; \*\mu^p, \*\Sigma_p)$ and
    \begin{IEEEeqnarray}{rClCrCl}
            (\*\Sigma^{\ast})^{-1} & =& \alpha\*\Sigma_q^{-1} + (1-\alpha)\*\Sigma_p^{-1}; & \quad &
        \*\mu^{\ast} & = & 
        \*\Sigma^{\ast}\left(\alpha\*\Sigma_q^{-1}\*\mu^q +
        (1-\alpha)\*\Sigma_p^{-1}\*\mu^p \right) \nonumber
    \end{IEEEeqnarray}
    it holds that for $\alpha \in (0,1)$, 
    \begin{IEEEeqnarray}{rCl}
        \RAD\left(q(\*\theta)||p(\*\theta)\right) &= &
        \dfrac{1}{2\alpha(1-\alpha)}\Big\{
            -\alpha\left[\*\mu^{q'}\*\Sigma_q^{-1}\*\mu^q + \ln|\*\Sigma_q| \right]
            -(1-\alpha)\left[\*\mu^{p'}\*\Sigma_p^{-1}\*\mu^p + \ln|\*\Sigma_p| \right] \nonumber \\
        && \;\; 
        \quad\quad\quad\quad\quad\quad
        + \left[ \*\mu^{\ast'}(\*\Sigma^{\ast})^{-1}\*\mu^{\ast} + \ln|\*\Sigma^{\ast}| \right]
        \Big\}
    \end{IEEEeqnarray}
Notice that computing this is of the same order as computing the \KLD uncertainty quantifier. In particular, one needs to perform a cholesky decomposition of $\*\Sigma_q$ and $\*\Sigma_q$ for either choice of $D$. 
\label{Thm:DivClosedForm}
\end{theorem}

\subsection{\GVI objectives for \DGP}

Using the layer-specific uncertainty quantifier $D^l\in\{\RAD, \frac{1}{w}\KLD\}$ for the \DGP layer with index $l$, the \GVI formulations of robust \DGPs within this report have the generalized form of eq. \eqref{eq:DGP_elbo_old} given by
\begin{IEEEeqnarray}{rCl}
    \mathcal{L}(q|\*Y, \*X, \ell, \{D^l\}_{l=1}^L) & = &
    \sum_{i=1}^n\mathbb{E}_{q(\*f^{L}_i)}
    \left[ 
        \ell(\*f^{L}_i, \*y_i)
    \right]
    +
    \sum_{l=1}^LD^l(q(\*F^l, \*U^l)||p(\*F^l, \*U^l |\*Z^{l-1})). \quad
    \label{eq:DGP_GVI_form1}
\end{IEEEeqnarray}
For the \DGP, Thms. \ref{Thm:closedFormExp} and \ref{Thm:DivClosedForm} show that the relevant quantities of this objective will be available in closed form.
Conceptually, the extension to generalized losses is straightforward \cite{Bissiri}. Moreover,  the same cannot be said about the new uncertainty quantifier. 
In particular, two important questions arise at this point: 
\begin{itemize}
    \item[(I)] 
    Will the divergence term simplify to 
    $\sum_{l=1}^LD^l(q(\*U^l)||p(\*U^l|\*Z^{l-1}))$ as in eq. \eqref{eq:DGP_elbo}?
    \item[(II)]
    Is $\sum_{l=1}^LD^l(q(\*F^l, \*U^l)||p( \*F^l, \*U^l|\*Z^{l-1}))$ a valid divergence between the full prior $\pi$ of eq. \eqref{eq:DGP_prior} and the variational posterior $q$ of eq. \eqref{eq:variationalFam1}?
\end{itemize}
%

\subsubsection{Does the layer-specific divergence define a valid divergence?}

To see that (I) can be answered positively, one simply needs to re-examine eq. (10) -- (12) in \citet{InducingBonilla}. 
In particular, note that for any divergence $D'(q||p)$ that can be written as $D'(q||p) = g(D(q||p))$ for some function $g(x)$ such that $g(x) \geq 0$ and $g(x) = 0$ if and only if $x=0$ and for some f-divergence $D^l(q||p) = \int_{\*F^l, \*U}q(\*F^l, \*U^l)f\left( \frac{q(\*F^l, \*U^l)}{p(\*F^l, \*U^l|\*Z^{l-1})}\right)d(\*F^l, \*U^l)$, it holds that
\begin{IEEEeqnarray}{rClCl}
    D'(q(\*F^l, \*U^l)||p(\*F^l, \*U^l|\*Z^{l-1}))
    & = &
    g \left(
    \mathbb{E}_{q(\*F^l, \*U^l)}\left[     f\left( \frac{q(\*F^l, \*U^l)}{p(\*F^l, \*U^l|\*Z^{l-1})}\right)
    \right]\right) \nonumber \\
    & = &
    g \left(
    \mathbb{E}_{p(\*F^l| \*U^l, \*Z^{l-1})q(\*U^l)}\left[     f\left( \frac{q(\*F^l, \*U^l)}{p(\*F^l, \*U^l|\*Z^{l-1})}\right)
    \right]\right) \nonumber \\
    & = &
    g \left(
    \mathbb{E}_{q(\*U)}\left[     f\left( \frac{q(\*U^l)}{p(\*U^l|\*Z^{l-1})}\right)
    \right]\right) \nonumber \\
    & = &
    D'(q(\*U^l)||p(\*U^l|\*Z^{l-1})).
\end{IEEEeqnarray}
This clearly holds for the special case of the \RAD with $g(x) = \frac{1}{\alpha(1-\alpha)}\log(x+1)$ and $f(x) = x^{1-\alpha}$.

\subsubsection{Does the layer-specific divergence simplify?}

The answer to (II)  is less obvious and
%
requires the following technical result  \citep[see][]{GVI}.
\begin{theorem}[Divergence recombination]
    {Let $D^l$ be divergences and  $c_l\geq0$ scalars for $l=1,2,\dots, L$. 
    Moreover, denote $\*\theta = (\*\theta_1, \*\theta_2, \dots, \*\theta_L)^T$, $\*\theta_{-l} = (\*\theta_1, \*\theta_2, \dots, \*\theta_{l-1}, \*\theta_{l+1},\dots \*\theta_L)^T$ and $\*\theta_{-(1:L)} = (\*\theta_{-1}, \*\theta_{-2}, \dots, \*\theta_{-L})^T$.
    Further, define the conditionally independent densities $p(\*\theta) = \prod_{l=1}^Lp_l(\*\theta_l|\*\theta_{-l})$ and $q(\*\theta) = \prod_{l=1}^Lq_l(\*\theta_l|\*\theta_{-l})$ and
    the conditioning-set dependent function $\widetilde{D}^{\*\theta'_{-(1:L)}}(q||p) = \sum_{l=1}^Lc_lD^l(q_l(\*\theta_l|\*\theta'_{-l})||p_l(\*\theta_l|\*\theta'_{-l}))$. 
    Then, $\widetilde{D}^{\*\theta_{-(1:L)}}(q||p)$ defines a divergence between $q$ and $p$ if (i) a version of the Hammersley-Clifford Theorem holds for $\{q_l, p_l:l = 1,2,\dots, L\}$ and (ii) $\widetilde{D}^{\*\theta'_{-(1:L)}}(q||p) = \widetilde{D}^{\*\theta^{\circ}_{-(1:L)}}(q||p)$ for any fixed pair of conditioning sets $\*\theta^{\circ}_{-(1:L)}, \*\theta'_{-(1:L)}$.}
    \label{Thm:DivRecomb}
\end{theorem}

Since conditions (i) and (ii) of this Theorem are easily verified for the \DGP as long as $D^l \in \{\RAD, \KLD\}$ for all $l=1,2,\dots, L$, the answer to (I) is also positive\footnote{see the supplementary material in \citep{GVI} for a  detailed verification of conditions (i) and (ii)}.

\begin{figure}[t!]
\vskip -0.5cm
\begin{center}
\includegraphics[trim= {2.2cm 0.2cm 2.85cm 0.1cm}, clip,  
width=1\columnwidth]{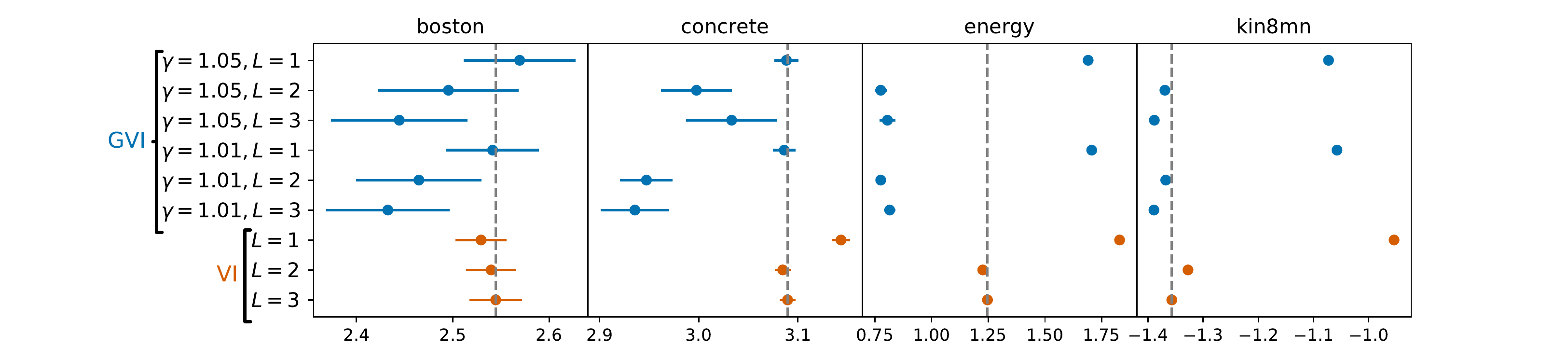}
\includegraphics[trim= {2.2cm 0.2cm 2.85cm 0.8cm}, clip,  
width=1\columnwidth]{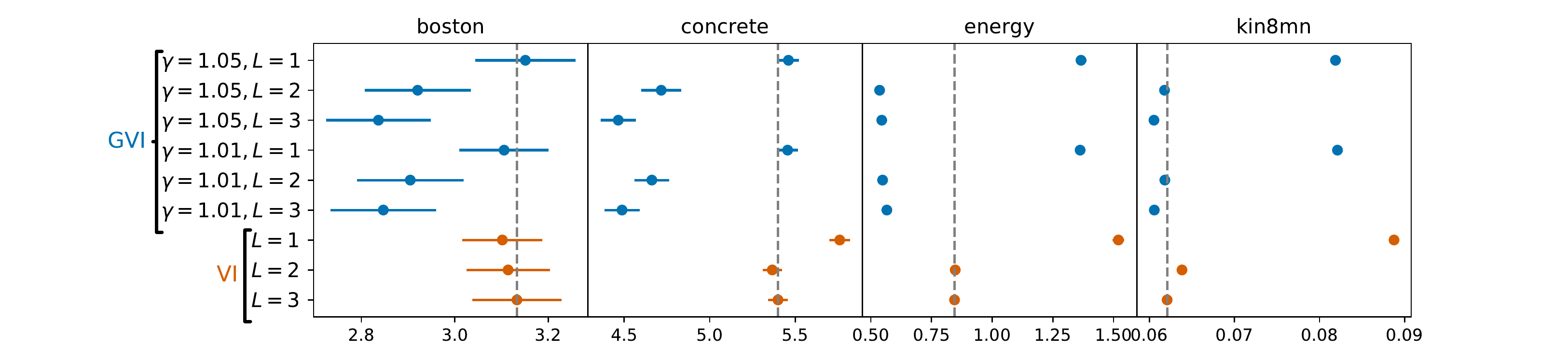}
\includegraphics[trim= {2.25cm 0.0cm 3cm 0.1cm}, clip,  
width=1\columnwidth]{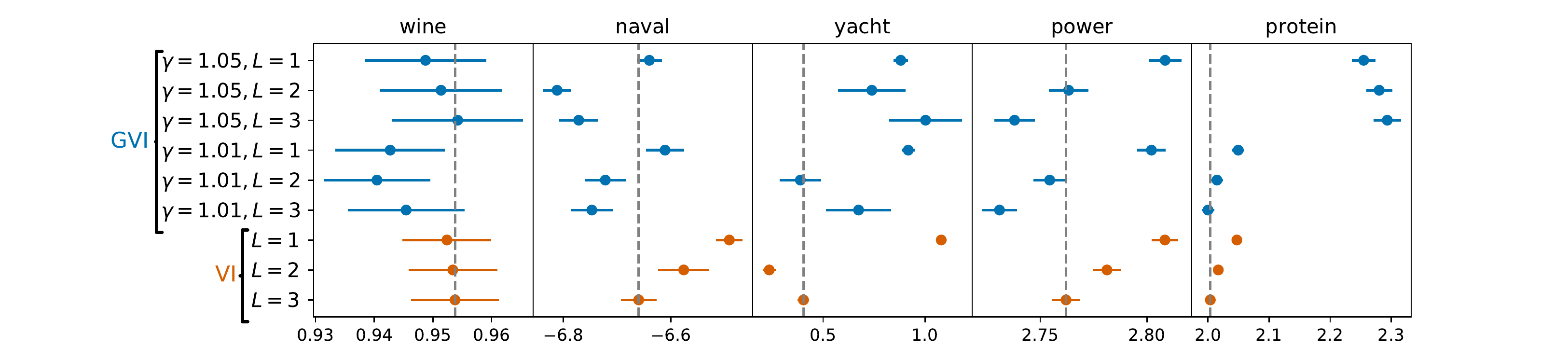}
\includegraphics[trim= {2.25cm 0.0cm 3cm 0.9cm}, clip,  
width=1\columnwidth]{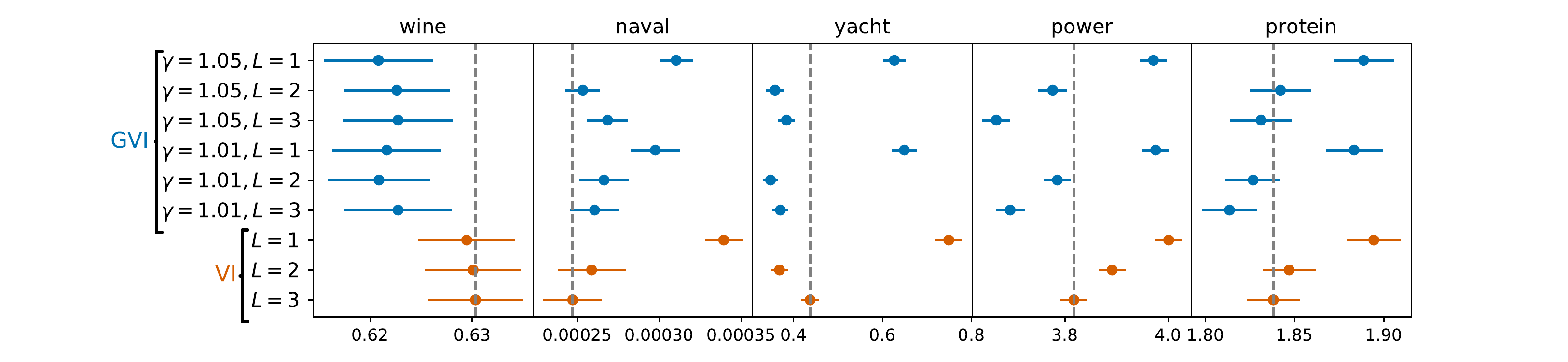}
\caption{
    Comparing performance in \DGP{}s with $L$ layers for 
    \textcolor{GVIColor1}{\textbf{\DGP-\GVI}} with 
    $\ell_n(\*\theta, \*x) = \sum_{i=1}^n\Lg(\*\theta, x_i)$ and \textcolor{VIColor}{\textbf{\DGP-\VI}}. Benchmark performance is the \DGP with three layers as in \citep{DeepGPsVI}. \textbf{Top rows}: Negative test log likelihoods. \textbf{Bottom rows}: Test \RMSE. The lower the better.
}
\label{Fig:DGPResults}
\vskip -0.5in
\end{center}
\end{figure}

\section{Experiments}

This section restates the experiments reported in \citet{GVI}. As in \citet{DeepGPsVI}, the method of choice is doubly stochastic (generalized) black box \VI/\GVI.  
Note that the supplementary material of \citep{GVI} contains more detail on the experimental setup as well as additional results. The code will be available publicly upon publication of \citet{GVI}\footnote{and will then be provided at \url{https://github.com/JeremiasKnoblauch/GVIPublic}}.

\textbf{Setup:}
Test set likelihoods and \RMSE{}s are reported by averaging over 50 random splits with 90\% training and 10\% test data. 
The \GVI methods provide robustness via $\Lg(\*\theta, x_i)$ rather than $\Lb(\*\theta, x_i)$ as $\Lg>0$, which allows for a log representation.
This is especially attractive on \DGP{}s due to the importance of numerical stability.
We use the variational family and code base of \citet{DeepGPsVI}. Except for choosing 50 (instead of 20) -fold cross validation with a 10\% randomly selected held out test set, all settings are the same as in \citet{DeepGPsVI}:
As in their paper, each experiment runs with ADAM \citep{ADAM} and a learning rate of 0.01 with 20,000 iterations.
%
For the kernel, we choose the RBF
kernel with a lengthscale for each dimension.
The number of inducing points is 100 for all settings, and they are run after normalization with a whitened representation.
The batch size is $\min(1000, n)$, where $n$ is the number of observations in the training set.
Each layer has $\min(30, D)$ latent functions, that is to say $D^l = \min(30, D)$ for all $l$.
The Python implementation extends the one of \citep{DeepGPsVI} and is based on tensorflow \citep{tensorflow} and gpflow \citep{gpflow}.

\begin{figure}[h!]
\vskip -0.5cm
\begin{center}
\includegraphics[trim= {0.0cm 0.5cm 2.85cm 0.1cm}, clip,  
width=1\columnwidth]{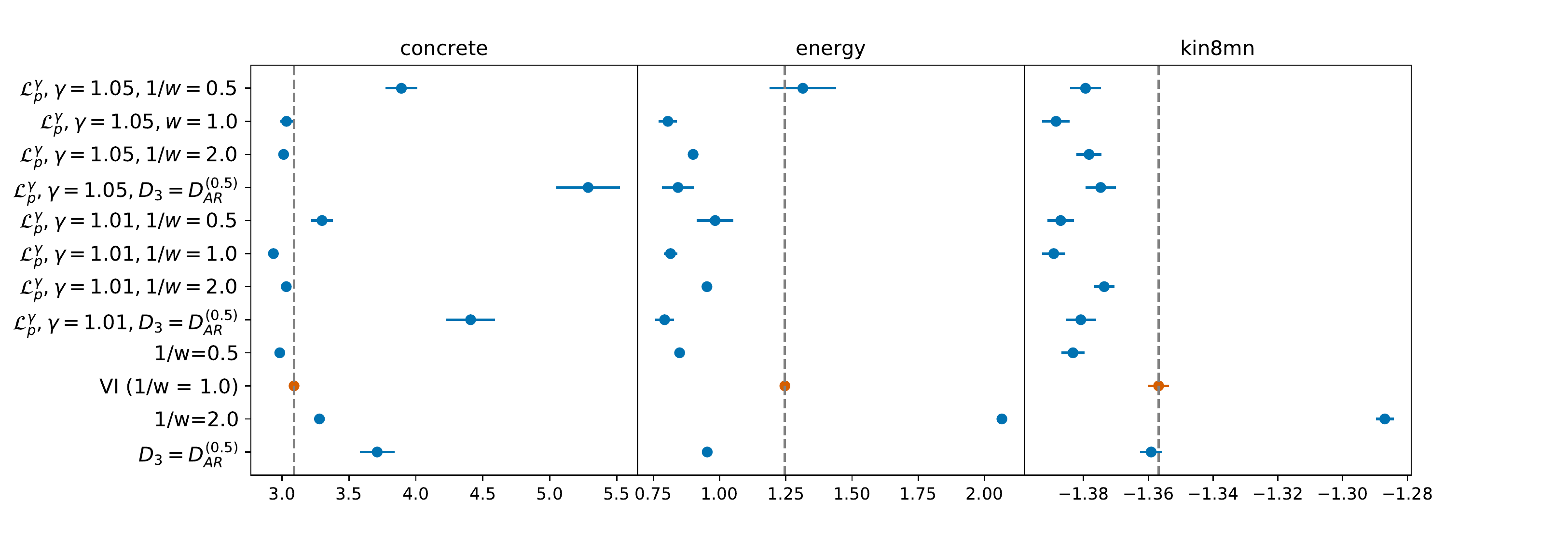}
\includegraphics[trim= {0.0cm 0.5cm 2.85cm 1.35cm}, clip,  
width=1\columnwidth]{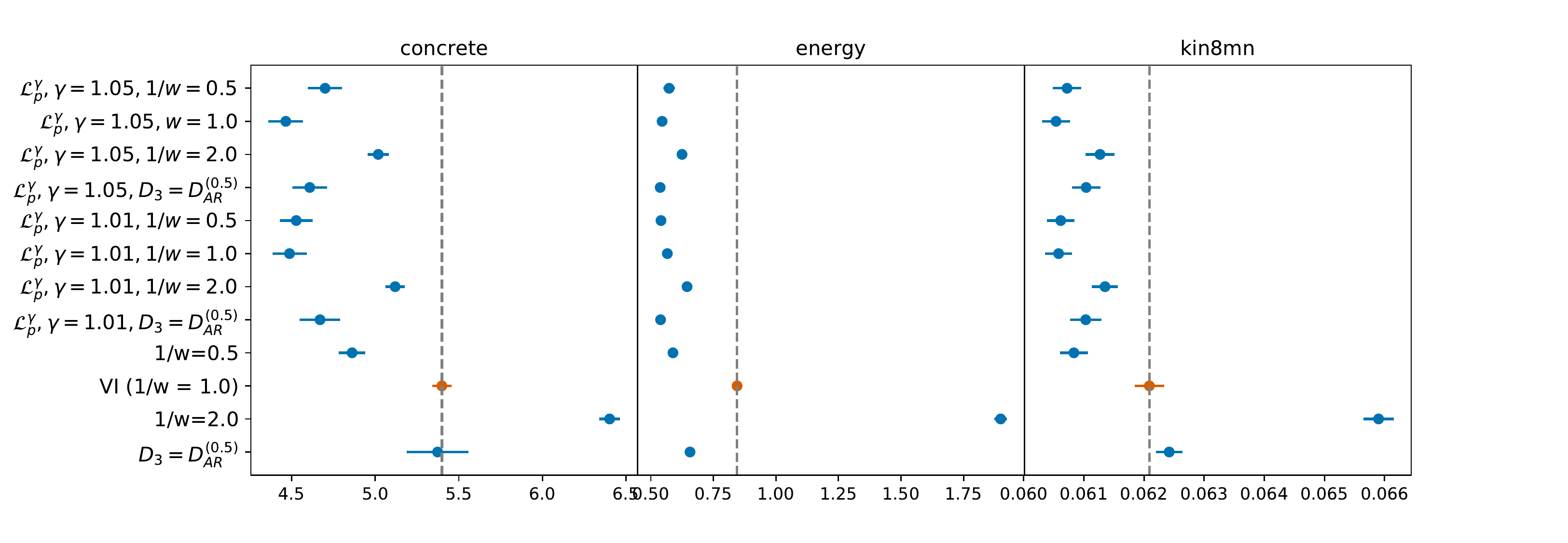}
\caption{
    Comparing performance in \DGP{}s with 3 layers for 
    \textcolor{GVIColor1}{\textbf{\DGP-\GVI}} with 
    $\ell_n(\*\theta, \*x) = \sum_{i=1}^n\Lg(\*\theta, x_i)$ and alternative uncertainty quantifiers $D$ against \textcolor{VIColor}{\textbf{\DGP-\VI}}.  Benchmark performance is the \DGP with three layers as in \citep{DeepGPsVI}. \textbf{Top row}: Negative test log likelihoods. \textbf{Bottom row}: Test \RMSE. The lower the better.
}
\label{Fig:DGPResults2}
\vskip -0.5in
\end{center}
\end{figure}

\textbf{Results \& Interpretation:}
The results for using $\Lg(\*\theta, x_i)$ are shown in Fig. \ref{Fig:DGPResults}. 
We find that \DGPs can benefit substantially from robust losses. 
This true for both the test \RMSE and the test likelihoods, giving a strong indication that information-geometric considerations are of considerable importance when dealing with large-scale, black-box type Bayesian models.
Moreover, Fig. \ref{Fig:DGPResults2} gives an overview over alternative uncertainty quantifiers $D$ as well as alternative losses $\Lg$. 
While the results indicate that similarly to Bayesian Neural Networks \citep[see][]{GVI}, enforcing more conservative uncertainty quantification typically does not lead to better test scores. 
In fact, the contrary is the case. This finding is explained by noting that any kind of deep architecture is likely to have an  extremely adaptable mean function. 
Thus, inflating the variance around this very informative mean function amounts to unwarranted uncertainty quantification.

\section*{Acknowledgements}

I most cordially thank Theodoros Damoulas and Jack Jewson for helpful comments and their ceaseless encouragements along the way.
Similarly, I thank Ollie Hamelijnck for useful suggestions for the final draft.
The funding making this work possible came from the \EPSRC grant EP/L016710/1 as well as from the Facebook Fellowship programme.
This work had additional support from the Lloyds Register Foundation programme on Data Centric Engineering through the London Air Quality project and was furthermore supported by The Alan Turing Institute for Data Science and AI under \EPSRC grant EP/N510129/1 in collaboration with the Greater London Authority.

\bibliography{library}

\begin{thebibliography}{32}
\providecommand{\natexlab}[1]{#1}
\providecommand{\url}[1]{\texttt{#1}}
\expandafter\ifx\csname urlstyle\endcsname\relax
  \providecommand{\doi}[1]{doi: #1}\else
  \providecommand{\doi}{doi: \begingroup \urlstyle{rm}\Url}\fi

\bibitem[Abadi et~al.(2016)Abadi, Barham, Chen, Chen, Davis, Dean, Devin,
  Ghemawat, Irving, Isard, et~al.]{tensorflow}
Mart{\'\i}n Abadi, Paul Barham, Jianmin Chen, Zhifeng Chen, Andy Davis, Jeffrey
  Dean, Matthieu Devin, Sanjay Ghemawat, Geoffrey Irving, Michael Isard, et~al.
\newblock Tensorflow: A system for large-scale machine learning.
\newblock In \emph{12th $\{$USENIX$\}$ Symposium on Operating Systems Design
  and Implementation ($\{$OSDI$\}$ 16)}, pages 265--283, 2016.

\bibitem[Basu et~al.(1998)Basu, Harris, Hjort, and Jones]{BasuDPD}
Ayanendranath Basu, Ian~R Harris, Nils~L Hjort, and MC~Jones.
\newblock Robust and efficient estimation by minimising a density power
  divergence.
\newblock \emph{Biometrika}, 85\penalty0 (3):\penalty0 549--559, 1998.

\bibitem[Bissiri et~al.(2016)Bissiri, Holmes, and Walker]{Bissiri}
Pier~Giovanni Bissiri, Chris~C Holmes, and Stephen~G Walker.
\newblock A general framework for updating belief distributions.
\newblock \emph{Journal of the Royal Statistical Society: Series B (Statistical
  Methodology)}, 78\penalty0 (5):\penalty0 1103--1130, 2016.

\bibitem[Bonilla et~al.(2016)Bonilla, Krauth, and Dezfouli]{InducingBonilla}
Edwin~V Bonilla, Karl Krauth, and Amir Dezfouli.
\newblock Generic inference in latent gaussian process models.
\newblock \emph{arXiv preprint arXiv:1609.00577}, 2016.

\bibitem[Bui et~al.(2016)Bui, Hern{\'a}ndez-Lobato, Hernandez-Lobato, Li, and
  Turner]{DGPEP}
Thang Bui, Daniel Hern{\'a}ndez-Lobato, Jose Hernandez-Lobato, Yingzhen Li, and
  Richard Turner.
\newblock Deep gaussian processes for regression using approximate expectation
  propagation.
\newblock In \emph{International Conference on Machine Learning}, pages
  1472--1481, 2016.

\bibitem[Cichocki and Amari(2010)]{ABCdiv}
Andrzej Cichocki and Shun-ichi Amari.
\newblock Families of alpha-beta-and gamma-divergences: Flexible and robust
  measures of similarities.
\newblock \emph{Entropy}, 12\penalty0 (6):\penalty0 1532--1568, 2010.

\bibitem[Cutajar et~al.(2017)Cutajar, Bonilla, Michiardi, and
  Filippone]{RandomFourierDGP}
Kurt Cutajar, Edwin~V Bonilla, Pietro Michiardi, and Maurizio Filippone.
\newblock Random feature expansions for deep gaussian processes.
\newblock In \emph{Proceedings of the 34th International Conference on Machine
  Learning-Volume 70}, pages 884--893. JMLR, 2017.

\bibitem[Dai et~al.(2015)Dai, Damianou, Gonz{\'a}lez, and Lawrence]{VAEDGPs}
Zhenwen Dai, Andreas Damianou, Javier Gonz{\'a}lez, and Neil Lawrence.
\newblock Variational auto-encoded deep gaussian processes.
\newblock \emph{arXiv preprint arXiv:1511.06455}, 2015.

\bibitem[Damianou and Lawrence(2013)]{DGPs}
Andreas Damianou and Neil Lawrence.
\newblock Deep gaussian processes.
\newblock In \emph{Artificial Intelligence and Statistics}, pages 207--215,
  2013.

\bibitem[Fujisawa and Eguchi(2008)]{GammaDivNotSummable}
Hironori Fujisawa and Shinto Eguchi.
\newblock Robust parameter estimation with a small bias against heavy
  contamination.
\newblock \emph{Journal of Multivariate Analysis}, 99\penalty0 (9):\penalty0
  2053--2081, 2008.

\bibitem[Futami et~al.(2018)Futami, Sato, and Sugiyama]{AISTATSBetaDiv}
Futoshi Futami, Issei Sato, and Masashi Sugiyama.
\newblock Variational inference based on robust divergences.
\newblock In \emph{International Conference on Artificial Intelligence and
  Statistics}, pages 813--822, 2018.

\bibitem[Ghosh and Basu(2016)]{GoshBasuPseudoPosterior}
Abhik Ghosh and Ayanendranath Basu.
\newblock Robust bayes estimation using the density power divergence.
\newblock \emph{Annals of the Institute of Statistical Mathematics},
  68\penalty0 (2):\penalty0 413--437, 2016.

\bibitem[Hensman and Lawrence(2014)]{NestedVariationalDGPs}
James Hensman and Neil~D Lawrence.
\newblock Nested variational compression in deep gaussian processes.
\newblock \emph{stat}, 1050:\penalty0 3, 2014.

\bibitem[Hensman et~al.(2013)Hensman, Fusi, and Lawrence]{GPforBigData}
James Hensman, Nicolo Fusi, and Neil~D Lawrence.
\newblock Gaussian processes for big data.
\newblock In \emph{Uncertainty in Artificial Intelligence}, page 282, 2013.

\bibitem[Hern{\'a}ndez-Lobato and Adams(2015)]{PBP}
Jos{\'e}~Miguel Hern{\'a}ndez-Lobato and Ryan Adams.
\newblock Probabilistic backpropagation for scalable learning of bayesian
  neural networks.
\newblock In \emph{International Conference on Machine Learning}, pages
  1861--1869, 2015.

\bibitem[Hooker and Vidyashankar(2014)]{MinDisparities}
Giles Hooker and Anand~N Vidyashankar.
\newblock Bayesian model robustness via disparities.
\newblock \emph{Test}, 23\penalty0 (3):\penalty0 556--584, 2014.

\bibitem[Hung et~al.(2018)Hung, Jou, and Huang]{GammaDivSummable}
Hung Hung, Zhi-Yu Jou, and Su-Yun Huang.
\newblock Robust mislabel logistic regression without modeling mislabel
  probabilities.
\newblock \emph{Biometrics}, 74\penalty0 (1):\penalty0 145--154, 2018.

\bibitem[Jewson et~al.(2018)Jewson, Smith, and Holmes]{Jewson}
Jack Jewson, Jim Smith, and Chris Holmes.
\newblock Principles of {B}ayesian inference using general divergence criteria.
\newblock \emph{Entropy}, 20\penalty0 (6):\penalty0 442, 2018.

\bibitem[Kingma and Ba(2014)]{ADAM}
Diederik~P Kingma and Jimmy Ba.
\newblock Adam: A method for stochastic optimization.
\newblock \emph{arXiv preprint arXiv:1412.6980}, 2014.

\bibitem[Kingma et~al.(2015)Kingma, Salimans, and
  Welling]{reparameterizationTrick}
Durk~P Kingma, Tim Salimans, and Max Welling.
\newblock Variational dropout and the local reparameterization trick.
\newblock In \emph{Advances in Neural Information Processing Systems}, pages
  2575--2583, 2015.

\bibitem[Knoblauch et~al.(2018)Knoblauch, Jewson, and Damoulas]{RBOCPD}
Jeremias Knoblauch, Jack Jewson, and Theodoros Damoulas.
\newblock Doubly robust {B}ayesian inference for non-stationary streaming data
  using $\beta$-divergences.
\newblock In \emph{Advances in {N}eural {I}nformation {P}rocessing {S}ystems
  ({N}eur{I}{P}{S})}, pages 64--75, 2018.

\bibitem[Knoblauch et~al.(2019)Knoblauch, Jewson, and Damoulas]{GVI}
Jeremias Knoblauch, Jack Jewson, and Theodoros Damoulas.
\newblock Generalized variational inference.
\newblock \emph{arXiv preprint arXiv:1904.02063}, 2019.

\bibitem[Kurtek and Bharath(2015)]{InfFct}
Sebastian Kurtek and Karthik Bharath.
\newblock Bayesian sensitivity analysis with the fisher--rao metric.
\newblock \emph{Biometrika}, 102\penalty0 (3):\penalty0 601--616, 2015.

\bibitem[Matthews et~al.(2016)Matthews, Hensman, Turner, and
  Ghahramani]{InducingProcesses}
Alexander G de~G Matthews, James Hensman, Richard Turner, and Zoubin
  Ghahramani.
\newblock On sparse variational methods and the kullback-leibler divergence
  between stochastic processes.
\newblock \emph{Journal of Machine Learning Research}, 51:\penalty0 231--239,
  2016.

\bibitem[Matthews et~al.(2017)Matthews, Alexander, Van Der~Wilk, Nickson,
  Fujii, Boukouvalas, Le{\'o}n-Villagr{\'a}, Ghahramani, and Hensman]{gpflow}
De~G Matthews, G~Alexander, Mark Van Der~Wilk, Tom Nickson, Keisuke Fujii,
  Alexis Boukouvalas, Pablo Le{\'o}n-Villagr{\'a}, Zoubin Ghahramani, and James
  Hensman.
\newblock Gpflow: A gaussian process library using tensorflow.
\newblock \emph{The Journal of Machine Learning Research}, 18\penalty0
  (1):\penalty0 1299--1304, 2017.

\bibitem[Rezende et~al.(2014)Rezende, Mohamed, and Wierstra]{gradTrick}
Danilo~Jimenez Rezende, Shakir Mohamed, and Daan Wierstra.
\newblock Stochastic backpropagation and approximate inference in deep
  generative models.
\newblock \emph{arXiv preprint arXiv:1401.4082}, 2014.

\bibitem[Salimbeni and Deisenroth(2017)]{DeepGPsVI}
Hugh Salimbeni and Marc Deisenroth.
\newblock Doubly stochastic variational inference for deep gaussian processes.
\newblock In \emph{Advances in Neural Information Processing Systems}, pages
  4588--4599, 2017.

\bibitem[Titsias(2009)]{InducingOriginal}
Michalis Titsias.
\newblock Variational learning of inducing variables in sparse gaussian
  processes.
\newblock In \emph{Artificial Intelligence and Statistics}, pages 567--574,
  2009.

\bibitem[Vafa(2016)]{SamplingDGPs}
Keyon Vafa.
\newblock Training deep gaussian processes with sampling.
\newblock In \emph{NIPS 2016 Workshop on Advances in Approximate Bayesian
  Inference}, 2016.

\bibitem[Wang et~al.(2016)Wang, Brubaker, Chaib-Draa, and Urtasun]{SMCDGPs}
Yali Wang, Marcus Brubaker, Brahim Chaib-Draa, and Raquel Urtasun.
\newblock Sequential inference for deep gaussian process.
\newblock In \emph{Artificial Intelligence and Statistics}, pages 694--703,
  2016.

\bibitem[Yang et~al.(2017)Yang, Pati, and Bhattacharya]{alpha-VI}
Yun Yang, Debdeep Pati, and Anirban Bhattacharya.
\newblock $\alpha$-variational inference with statistical guarantees.
\newblock \emph{arXiv preprint arXiv:1710.03266}, 2017.

\bibitem[Zellner(1988)]{Zellner}
Arnold Zellner.
\newblock Optimal information processing and bayes's theorem.
\newblock \emph{The American Statistician}, 42\penalty0 (4):\penalty0 278--280,
  1988.

\end{thebibliography}
\bibliographystyle{plainnat}

\end{document}